# Playing games against nature: optimal policies for renewable resource allocation

**Stefano Ermon**
Department of Computer Science
Cornell University
ermonste@cs.cornell.edu

**Jon Conrad**
Department of Applied Economics
Cornell University
jmc16@cornell.edu

**Carla Gomes, Bart Selman**
Department of Computer Science
Cornell University
{gomes,selman}@cs.cornell.edu

## Abstract

In this paper we introduce a class of Markov decision processes that arise as a natural model for many renewable resource allocation problems. Upon extending results from the inventory control literature, we prove that they admit a closed form solution and we show how to exploit this structure to speed up its computation.

We consider the application of the proposed framework to several problems arising in very different domains, and as part of the ongoing effort in the emerging field of Computational Sustainability we discuss in detail its application to the Northern Pacific Halibut marine fishery. Our approach is applied to a model based on real world data, obtaining a policy with a guaranteed lower bound on the utility function that is structurally very different from the one currently employed.

## 1 Introduction

The problem of devising policies to optimally allocate resources over time is a fundamental decision theoretic problem with applications arising in many different fields. In fact, such decisions may involve a variety of different resources such as time, energy, natural and financial resources, in allocation problems arising in domains as diverse as natural resources management, crowdsourcing, supply chain management, QoS and routing in networks, vaccine distribution and pollution management.

A particularly interesting class of such problems involves policies for the allocation of *renewable resources*. A key and unique aspect of such a resource type is the fact that, by definition, its stock is constantly replenished by an intrinsic growth process. The most common example are perhaps living resources, such as fish populations or forests, that increase constantly by natural growth and reproduction, but less conventional resources such as users in a social community or in a crowdsourcing project share the same intrinsic growth feature due to social interactions.

A common feature of the growth processes presented is that they are density dependent, in the sense that the growth rate depends on the amount of resource available. This fact creates a challenging management problem when the aim of the intervention is to optimally use the resource, for instance by harvesting a fish population or by requiring some effort from a crowdsourcing community, especially when economic aspects are factored in. We face a similar challenge in vaccine distribution problems, where the growth rate of infections is again density dependent and the objective is to reduce its spreading.

This study, in particular, has been motivated by the alarming consideration that many natural resources are endangered due to over-exploitation and generally poorly managed. For instance, the Food and Agricultural Organization estimates in their most recent report that $7\%$ of marine fish stocks are already depleted, $1\%$ are recovering from depletion, $52\%$ are fully exploited and $17\%$ are overexploited ([1]).

One of the most fundamental aspects of the problem seems to be the lack of an effective way to handle the uncertainty affecting the complex dynamics involved. While in most of the works in the literature [6, 7] these growth processes are modeled with deterministic first-order difference or differential equations, this approach often represents an oversimplification. In fact their intrinsic growth is often affected by many variables and unpredictable factors. For example, in the case of animal populations such as fisheries, both weather and climate conditions are known to affect both the growth and the mortality in the population. Other variable ecological factors such as the availability of food or the interaction with other species also influence their natural dynamics to the point that it is very difficult even to obtain reliable mathematical models to describe their dynamics.

On the other hand, stochastic differential equations can easily incorporate these variable factors and therefore represent a more robust description. However, obtaining a prob-abilistic description of such systems is far from easy. In fact, even if in principle uncertainty could be reduced by collecting and analyzing more data, it is generally believed that complex and stochastic systems, such a marine environments, could never become predictable (to the point that the authors of [13] believe that "predictability of anything as complex as marine ecosystem will forever remain a chimera").

Moreover, there are situations of "radical uncertainty" ([8]) or ambiguity where a stochastic description is not feasible because the probabilities are not quantifiable. For instance, many fundamental environmental issues that we are facing, such as those surrounding the climate change debate, involve ambiguity in the sense of scientific controversies or irreducible beliefs that cannot be resolved.

In the context of stochastic optimization, there are two main ways to deal with uncertainty. The first one involves a *risk management* approach, where it is assumed that the probabilities of the stochastic events are known a priori or are learned from experience through statistical data analysis. Within this framework, decisions are taken according to stochastic control methods. Using tools such as risk-sensitive Markov decision processes ([12, 15]), it is also possible to encode into the problem the attitude towards risk of the decision maker by using an appropriate *utility function*. In particular the degree of risk aversion can be controlled by sufficiently penalizing undesirable outcomes with the utility function. When a fine grained stochastic description is not available, worst-case game theoretic frameworks, that are inherently risk averse, play a fundamental role because it is often crucial to devise policies that avoid catastrophic depletion. This type of approach, where the problem of data uncertainty is addressed by guaranteeing the optimality of the solution for the worst realizations of the parameters, is also known in the literature as *robust optimization* ([3, 5]), and has been successfully applied to uncertain linear, conic quadratic and semidefinite programming.

In this paper, we present a class of Markov decision processes that arise as a natural model for many resource management problems. Instead of formulating the optimization problem in a traditional form as a maximization of an *expected* utility, we tackle the management problems in a game theoretic framework, where the optimization problem is equivalent to a *dynamic game against nature*. This formulation is a particular type of *Markov game* [14] (sometimes called a *stochastic game* [16]) where there are only two agents (the manager and nature) and they have diametrically opposed goals.

As mentioned before, although this formulation is more conservative, it also eliminates the very difficult task of estimating the probabilities of the stochastic events affecting the system. In a context where the emphasis in the literature has traditionally been on the study of expected utilities, this approach represents a new perspective. Moreover, the policies thus obtained provide a lower bound on the utility that can be guaranteed to be achieved, no matter the outcomes of the stochastic events. For this class of problems, we are able to completely characterize the optimal policy with a theoretical analysis that extends results from the inventory control literature, obtaining a closed form solution for the optimal policy.

As part of the new exciting research area of Computational Sustainability ([10]), where techniques from computer science and related fields are applied to solve the pressing sustainability challenges of our time, we present an application of the proposed framework to the Northern Pacific Halibut fishery, one of the largest and most lucrative fisheries of the Northwestern coast. In particular, our method suggests the use of a cyclic scheme that involves periodic closures of the fishery, a policy that is structurally different from the one usually employed, that instead tries to maintain the stock at a given size with appropriate yearly harvests. However, this framework is interesting in its own right and, as briefly mentioned before, it applies to a variety of other problems that share a similar mathematical structure and that arise in very different domains. For example, we can apply our framework to pollution problems, where a stock of pollutants is evolving over time due to human action, and the objective is to minimize the total costs deriving from the presence of a certain stock of pollutants and the costs incurred with cleanups, but also to crowdsourcing and other problems.

## 2 MDP Formulation

In this section, we will formulate the optimization problem as discrete time, continuous space Markov decision process. Whenever possible, we will use a notation consistent with the one used in [4]. Even if we will consider only a finite horizon problem, the results can be extended to the infinite horizon case with limiting arguments. To make the description concrete, the model will be mostly described having a natural resource management problem in mind.

We consider a dynamical system evolving over time according to

$$x_{n+1} = f(x_n - h_n, w_n), \quad (1)$$

where $x_n \in \mathbb{R}$ denotes the stock of a renewable resource at time $n$. By using a discrete time model we implicitly assume that replacement or birth processes occur in regular, well defined "breeding seasons", where $f(\cdot)$ is a *reproduction function* that maps the stock level at the end of one season to the new stock level level at the beginning of the next season. The control or decision variable at year $n$ is the harvest level $h_n$ (occurring between two consecutive breeding seasons), that must satisfy $0 \leq h_n \leq x_n$.

As mentioned in the introduction, the function $f(\cdot)$ cap-tures the intrinsic replenishment ability of renewable resources, that in many practical applications (such as fisheries or forestry) is density dependent: growth rate is high when the habitat is underutilized but it decreases when the stock is larger and intraspecific competition intensifies. Specific properties of reproduction functions $f(\cdot)$ will be discussed in detail later, but we will always assume that there is a finite maximum stock level denoted by $m$.

To compensate for the higher level description of the complex biological process we are modeling, we introduce uncertainty into the model through $w_n$, a random variable that might capture, for example, the temperature of the water, an uncontrollable factor that influences the growth of the resource. Given the worst case framework we are considering, we will never make assumptions on the probability distribution of $w_n$ but only on its support (or, in other words, on the possible outcomes). In fact in an adversarial setting it is sufficient to consider all possible scenarios, each one corresponding to an action that nature can take against the policy maker, without assigning them a weight in a probabilistic sense.

Given the presence of stochasticity, it is convenient to consider closed loop optimization approaches, where decisions are made in stages and the manager is allowed to gather information about the system between stages. In particular, we assume that the state of the system $x_n \in \mathbb{R}$ is completely observable. For example, in the context of fisheries this means that we assume to know exactly the level of the stock $x_n$ when the harvest level $h_n$ is to be chosen. In this context, a *policy* is a sequence of rules used to select at each period a harvest level for each possible stock size. In particular, an *admissible policy* $\pi = \{\mu_1, \ldots, \mu_N\}$ is a sequence of functions, each one mapping stocks sizes $x$ to harvests $h$, so that for all $x$ and for all $i$

$$0 \leq \mu_i(x) \leq x. \quad (2)$$

## 2.1 Resource Economics

We now consider the economic aspects of the model. We suppose that the revenue obtained from a harvest $h$ is proportional to $h$ through a fixed price $p$, and that harvesting is costly. In particular we assume that there is

- a fixed set-up cost $K$ each time a harvest is undertaken
- a marginal harvest cost $g(x)$ per unit harvested when the stock size is $x$

It follows that the utility derived from a harvest $h$ from an initial stock $x$ is

$$ph - \int_{x-h}^{x} g(y)dy - K \triangleq R(x) - R(x-h) - K, \quad (3)$$

where

$$R(x) = px - \int_0^x g(y)dy.$$

We assume that the marginal harvesting cost $g(x)$ increases as the stock size $x$ decreases. We include time preference into the model by considering a fixed discount factor $\alpha = 1/(1+\delta)$ ( $0 \leq \alpha \leq 1$), where $\delta > 0$ is a *discount rate*.

For any given horizon length $N$, we consider the problem of finding an *admissible policy* $\pi = \{\mu_i\}_{i \in [1,N]}$ that maximizes

$$C_N^\pi(x) = \min_{\substack{w_1, \ldots, w_N \\ w_i \in W(x_i)}} \sum_{n=1}^{N} \alpha^n (R(x_n) - R(x_n - h_n) - K\delta_0(h_n))$$

where $x_n$ is subject to (1) and $h_n = \mu_n(x_n)$, with initial condition $x_1 = x$ and

$$\delta_0(x) = \begin{cases} 1 & \text{if } x > 0, \\ 0 & \text{otherwise.} \end{cases}$$

This is a Max-Min formulation of the optimization problem, where the goal is to optimize the utility in a worst-case scenario. As opposed to the maximization of an *expected* utility ([17, 18]), this formulation is inherently risk averse. An advantage of this formulation is that there is no need to characterize the probability distribution of the random variables $w_k$ explicitly, but only to determine their support. In fact, one should consider all the possible scenarios, without worrying about the probabilities of their occurrence.

## 3 Main Results

### 3.1 Minimax Dynamic Programming

A policy $\pi$ is called an optimal $N$-period policy if $C_N^\pi(x)$ attains its supremum over all admissible policies at $\pi$ for all $x$. We call

$$C_N(x) = \sup_{\pi \in \Pi} C_N^\pi(x),$$

the *optimal value function*, where $\Pi$ represents the set of all admissible policies.

As a consequence of the principle of optimality([4]), the dynamic programming equation for this problem reads:

$$\begin{aligned} C_0(x) &= 0, \\ C_n(x) &= \max_{0 \leq h_n \leq x} \min_{w_n \in W} R(x_n) - R(x_n - h_n) \\ &\quad - K\delta_0(h_n) + \alpha C_{n-1}(f(x - h_n, w_n)) \end{aligned}$$

for all $n > 0$. The latter equation can be rewritten in terms of the remaining stock $z = x - h_n$ (the post decision state) as

$$C_n(x) = \alpha \max_{0 \leq z \leq x} \left( R(x) - R(z) - K\delta_0(x - z) + \min_{w_n \in W} C_{n-1}(f(z, w_n)) \right). \quad (4)$$

This formulation of the problem is effectively analogous to a *game against nature* in the context of a two-person zero-sum game. The objective is in fact devising the value of $z$ that maximizes the utility, but assuming that nature is actively playing against the manager with the opposite intention.

It can be shown (see [4]) that $C_n(x)$, the revenue function associated with an optimal policy, is the (unique) solution to equation (4). From equation (4) we see that an optimal policy, when there are $n$ periods left and the stock level is $x$, undertakes a harvest if and only if there exists $0 \leq z \leq x$ such that

$$R(x) - R(z) - K + \alpha \min_{w_n \in W} C_{n-1}(f(z,w_n)) > \alpha \min_{w_n \in W} C_{n-1}(f(x,w_n)).$$

In fact, an action should be taken if and only if its associated benefits are sufficient to compensate the fixed cost incurred. By defining

$$P_n(x) = -R(x) + \alpha \min_{w_n \in W} C_{n-1}(f(x,w_n)), \quad (5)$$

we have that an optimal policy, when there are $n$ periods left and the stock level is $x$, undertakes a harvest if and only if there exists $0 \leq z \leq x$ such that

$$P_n(z) - K > P_n(x). \quad (6)$$

To examine this kind of relationship it is useful to introduce the notion of $K$-concavity, a natural extension of the $K$-convexity property originally introduced by Scarf in [19] to study inventory control problems.

### 3.2 Preliminaries on K-concavity

A function $\beta(\cdot)$ is $K$-concave if given three points $x < y < z$, $\beta(y)$ exceeds the secant approximation to $\beta(y)$ obtained using the points $\beta(x) - K$ and $\beta(z)$. Therefore for $K = 0$ no slack is allowed and one recovers the standard definition of concavity. Formally

**Definition 1.** *A real valued function $\beta(\cdot)$ is $K$-concave if for all $x, y$, $x < y$, and for all $b > 0$*

$$\beta(x) - \beta(y) - (x-y)\frac{\beta(y+b) - \beta(y)}{b} \leq K. \quad (7)$$

We state some useful results concerning $K$-concavity:

**Lemma 1.** *The following properties hold:*

- *A concave function is $0$-concave and hence $K$-concave for all $K \geq 0$.*

- *If $\beta_1(q)$ and $\beta_2(q)$ are respectively $K_1$-concave and $K_2$-concave for constants $K_1 \geq 0$ and $K_2 \geq 0$, then $a\beta_1(q) + b\beta_2(q)$ is $(aK_1 + bK_2)$-concave for any scalars $a > 0$ and $b > 0$.*

- *If $\beta(\cdot)$ is nondecreasing and concave on $I$ and $\psi(\cdot)$ is nondecreasing and $K$-concave on $[\inf_{x \in I} \beta(x), \sup_{x \in I} \beta(x)]$ then the composition $\psi \circ \beta$ is $K$-concave on $I$.*

- *Let $\beta_1(x), \ldots, \beta_N(x)$ be a family of functions such that $\beta_i(x)$ is $K_i$-concave. Then $\gamma(x) = \min_i \beta_i(x)$ is $(\max_i K_i)$-concave.*

- *If $\beta(\cdot)$ is a continuous, $K$-concave function on the interval $[0, m]$, then there exists scalars $0 \leq S \leq s \leq m$ such that*

    - *$\beta(S) \geq \beta(q)$ for all $q \in [0, m]$.*
    - *Either $s = m$ and $\beta(S) - K \leq \beta(m)$ or $s < m$ and $\beta(S) - K = \beta(s) \geq \beta(q)$ for all $q \in [s, m)$.*
    - *$\beta(\cdot)$ is a decreasing function on $[s, m]$.*
    - *For all $x \leq y \leq s$, $\beta(x) - K \leq \beta(y)$.*

The proof is not reported here for space reasons, but can be found in [9]. Similar results for $K$-convex functions are proved in [4].

In the following section we will prove by induction the $K$-concavity of the functions $P_n(x)$, $n = 1, \ldots, N$. This will allow us to characterize the structure of the optimal policy by using the last assertion of Lemma 1.

### 3.3 On the Optimality of $(S - s)$ policies

Suppose that we can prove that $P_n(x)$ is continuous and strictly K-concave. Then by Lemma 1 there exists $S_n$, $s_n$ with the properties proved in the last point of the Lemma. It is easy to see that condition (6) is satisfied only if $x > s$, in which case the optimal value of the remaining stock $z$ would be precisely $S_n$. In conclusion, if we can prove the continuity and $K$-concavity of the functions $P_n(x)$, $n = 1, \ldots, N$, then following feedback control law, known as a nonstationary $(S - s)$ policy, is optimal:

At period $n$, a harvest is undertaken if and only if the current stock level is greater than $s_n$; in that case the stock is harvested down to $S_n$.

This policy is known in the inventory control literature as a nonstationary $(S-s)$ policy [1], because the levels $S_n$ and $s_n$ are time dependent. Since it is assumed that the marginal harvest cost $g(x)$ is a non increasing function, we define $x_0$ to be the zero profit level such that $g(x_0) = p$. If $g(x) < p$ for all $x$, we define $x_0 = 0$. As a consequence for all $x > x_0$ we have that $R'(x) \geq 0$ so that $R$ (defined in equation (3)) is non decreasing. Moreover if the marginal harvest cost $g(x)$ is a non increasing function, then $R$ is convex.

---

[1] For the sake of consistency, we call $s_n$ the threshold value that governs the decision, even if in our case $S_n \leq s_n$.

We also need to make an assumption on the concavity of $R(\cdot)$. In particular the marginal cost function $g$ is allowed to decrease but not by too much. Let $m$ be an upper bound on the possible values of $x$ and $G(x) = \int_0^x g(t)dt$, then we need

$$\tau = G(m) - mg(m) < K\left(\frac{1-\alpha}{\alpha}\right), \qquad (8)$$

a condition that implies the $\tau$-concavity of $R$.

The main result is the following theorem, where we show that if some assumptions are satisfied, the optimal policy is of $(S-s)$ type. The key point of this inductive proof is to show that the $K$-concavity property is preserved by the Dynamic Programming operator.

**Theorem 1.** *For any setup cost $K > 0$ and any positive integer $N$, if $f(\cdot, w)$ is nondecreasing and concave for any $w$ and if $g$ is non increasing and satisfies condition (8), then the functions $P_n(x)$ defined as in (5) are continuous and $K$-concave for all $n = 1, \ldots, N$. Hence there exists a non-stationary $(S-s)$ policy that is optimal. The resulting optimal present value functions $C_n(x)$ are continuous, nondecreasing and $K$-concave for all $n = 1, \ldots, N$.*

*Proof.* From equation (8) we know that there exists a number $k$ such that

$$(K+\tau)\alpha < k < K. \qquad (9)$$

The proof is by induction on $N$. The base case $N = 0$ is trivial because $C_0(x) = 0$ for all $x$, and therefore it is continuous, nondecreasing and $k$-concave. Now we assume that $C_n(x)$ is continuous, nondecreasing and $k$-concave, and we show that $P_{n+1}(x)$ is continuous and $K$-concave, and that $C_{n+1}(x)$ is continuous, nondecreasing and $k$-concave.

Since $f(\cdot, w)$ is nondecreasing and concave for all $w$, $C_n(f(z, w_n))$ is $K$-concave by Lemma (1). By Lemma 1

$$\min_{w_n \in W} C_{n-1}(f(z, w_n))$$

is also $K$-concave. Again using Lemma 1, if $-R(x)$ is concave, then by equation (5) $P_{n+1}(x)$ is $K$-concave. The continuity of $P_{n+1}(x)$ is implied by the continuity of $C_n(x)$ and $R(x)$.

Given that $P_{n+1}(x)$ is $K$-concave and continuous, the optimal action is to harvest down to $S_{n+1}$ if and only if the current stock level is greater than $s_{n+1}$, so we have

$$C_{n+1}(x) = \begin{cases} \alpha(P_{n+1}(x) + R(x)) & \text{if } x \leq s_{n+1}, \\ \alpha(P_{n+1}(S_{n+1}) + R(x) - K) & \text{if } x > s_{n+1}. \end{cases} \qquad (10)$$

The continuity of $C_{n+1}(x)$ descends from the continuity of $P_{n+1}(x)$ and because by definition $P_{n+1}(s_{n+1}) + R(s_{n+1}) = P_{n+1}(S_{n+1}) + R(s_{n+1}) - K$. To show it is nondecreasing, consider the case $0 \leq x_1 < x_2 \leq s_{n+1}$:

$$C_{n+1}(x_2) - C_{n+1}(x_1) =$$
$$\alpha\left(\min_{w_n \in W} C_n(f(x_2, w_n)) - \min_{w_n \in W} C_n(f(x_1, w_n))\right).$$

If for all $x_2 > x_1 \geq 0$,

$$\min_{w_n \in W(x_2)} f(x_2, w_n) \geq \min_{w_n \in W(x_1)} f(x_1, w_n),$$

then $C_{n+1}(x_2) - C_{n+1}(x_1) \geq 0$ because $C_n(x)$ is nondecreasing. For the case $s_{n+1} < x_1 < x_2$ and $s_{n+1} \geq x_0$:

$$C_{n+1}(x_2) - C_{n+1}(x_1) = \alpha(R(x_2) - R(x_1)) \geq 0,$$

because $R$ is nondecreasing on that interval. It must be the case that $S_{n+1} > x_0$ because harvesting below $x_0$ is not profitable and reduces the marginal growth of the stock, so given that $s_{n+1} \geq S_{n+1} \geq x_0$ we conclude that $C_{n+1}(x)$ is nondecreasing. It remains to show that $C_{n+1}(x)$ is $k$-concave, and by equation (9) it is sufficient to show that it is $(K+\tau)\alpha$-concave. To show that definition (7) holds for $C_{n+1}(x)$, we consider several cases.

When $x < y \leq s_{n+1}$, according to equation (10) we have that $C_{n+1}(x) = \alpha(P_{n+1}(x) + R(x))$ and therefore equation (7) holds by Lemma 1 because $P_{n+1}$ is $K$-concave and $R(\cdot)$ is $\tau$-concave. Similarly when $s_{n+1} < x < y$, equation (7) holds because $R(\cdot)$ is $\tau$-concave.

When $x \leq s_{n+1} < y$ equation (7) reads

$$C_{n+1}(x) - C_{n+1}(y) - (x-y)\frac{C_{n+1}(y+b) - C_{n+1}(y)}{b} \leq$$
$$\alpha\left(K + R(x) - R(y) - (x-y)\frac{R(y+b) - R(y)}{b}\right) \leq$$
$$\alpha(K+\tau).$$

because $P_{n+1}(x) \leq P_{n+1}(S_{n+1})$ and $R(\cdot)$ is $\tau$-concave. $\square$

## 4 Consistency and Complexity

Even if Theorem 1 completely describes the structure of the optimal policy, in general there is no closed form solution for the values of $S_n$ and $s_n$, that need to be computed numerically. In order to use the standard dynamic programming approach, the state, control and disturbance spaces must be discretized, for instance using an evenly spaced grid. Since we are assuming that those spaces are bounded, we obtain in this way discretized sets with a finite number of elements. We can then write DP like equations for those points, using an interpolation of the value function for the points that are not on the grid. The equations can be then solved recursively, obtaining the semi-optimal action to be taken for each point of the grid, that can then be extended by interpolation to obtain an approximate solution to the original problem.

As with all discretization schemes, we need to discuss the *consistency* of the method. In particular, we would like (uniform) convergence to the solution of the original problem in the limit as the discretization becomes finer. It is well known that in general this property does not hold. However in this case Theorem 1 guarantees the continuity of $C_n$, that in turn implies the *consistency* of the method, even if the policy itself is not continuous as a function of the state([4]). Intuitively, discrepancies are possible only around the threshold $s_n$, so that they tend to disappear as the discretization becomes finer.

The standard dynamic programming algorithm involves $O(|X||W||U||T|)$ arithmetic operations, where $|X|$ is the number of discretized states, $|W|$ the number of possible outcomes of the (discretized) uncontrollable events, $|U|$ the maximum number of possible discretized actions that can be taken in any given state and $T$ is the length of the time horizon. However, the priori knowledge of the structure of the optimal policy can be used to speed up the computation. In fact it is sufficient to find $s$ (for example by bisection) and compute the optimal control associated with any state larger than $s$ to completely characterize the policy for a given time step. The complexity of this latter algorithm is $O(|W||U||T| \log |X|)$.

## 5 Case Study: the Pacific Halibut

As part of the ongoing effort in the emerging field of Computational Sustainability, we consider an application of our framework to the Pacific Halibut fishery.
The commercial exploitation of the Pacific halibut on the Northwestern coastline of North America dates back to the late 1800s, and it is today one of the region's largest and most profitable fisheries. The fishery developed so quickly that by the early 20th century it was starting to exhibit signs of overfishing. After the publication of scientific reports which demonstrated conclusively a sharp decline of the stocks, governments of the U.S. and Canada signed a treaty creating the International Pacific Halibut Commission (IPHC) to rationally manage the resource. The IPHC commission controls the amount of fish caught annually by deciding each year's *total allowable catch* (TAC), that is precisely the decision variable $h_n$ of our optimization problem.

### 5.1 Management Problem Formulation

To develop a bioeconomic model of the fishery, we have extracted data [2] from the IPHC annual reports on estimated biomass $x_t$, harvest $h_t$ and effort $E_t$ (measured in thousands of skate soaks) for Area 3A (one of the major regulatory areas in which waters are divided) for a 33 years period from 1975 to 2007. To model the population dynamics, we

---

[2]Data is available from the authors upon request.

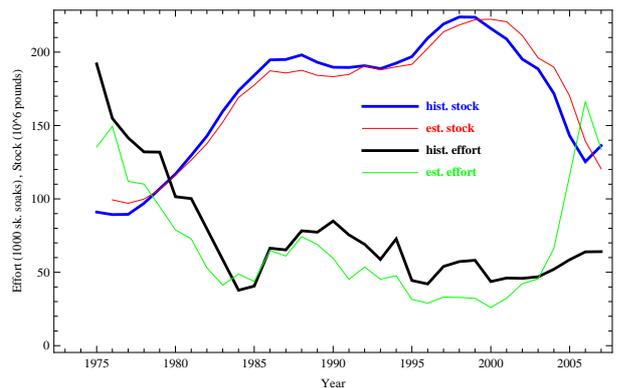

Figure 1: Fitted models (11) and (13) compared to historical data (in bold).

consider the Beverton-Holt model that uses the following reproduction function

$$x_{n+1} = f(s_n) = (1-m)s_n + \frac{r_0 s_n}{1 + s_n/M}, \quad (11)$$

where $s_n = x_n - h_n$ is the stock remaining after fishing (escapement) in year $n$. This model can be considered as a discretization of the continuous-time logistic equation. Here, parameter $m$ represents a natural mortality coefficient, $r_0$ can be interpreted as a reproduction rate and $M(r_0-m)/m$ is the carrying capacity of the environment. The (a priori) mortality coefficient we use is $m = 0.15$, that is the current working value used by the IPHC. The values of $r_0$ and $M$ are estimated by ordinary least square fitting to the historical data. Estimated values thus obtained are reported in table 1, while the fitted curve is shown in figure 1.

| Parameter | Value |
|---|---|
| $q$ | $9.07979 \; 10^{-7}$ |
| $b$ | $2.55465$ |
| $p$ | $4,300,000\$ \; / \; (10^6 \text{ pounds})$ |
| $K$ | $5,000,000\$$ |
| $c$ | $200,000\$ \; / \; 1000 \text{ skate soaks}$ |
| $\delta$ | $0.05$ |
| $m$ | $0.15$ |
| $M$ | $196.3923 \; 10^6 \text{ pounds}$ |
| $r_0$ | $0.543365$ |

Table 1: Base case parameter set.

Following [18], we suppose that the system is affected by stochasticity in the form of seasonal shocks $w_n$ that influence only the new recruitment part

$$x_{n+1} = f(s_n, w_n) = (1-m)s_n + w_n \frac{r_0 s_n}{1 + s_n/M}. \quad (12)$$

Instead of assuming an a priori probability distribution for $w_n$ or trying to learn one from data (that in our case would not be feasible given current scarce data availability), we will make use of the framework developed in the previous sections. In particular we will (a priori) assume that $w_n$ are random variables all having the same finite support that we will learn from data, but we will not make any assumption on the actual weight distribution. With our data, we obtain that $w_n \in [1 - 0.11, 1 + 0.06] = I_w$.

For the economic part of the model, we start by modeling the relationship between a harvest $h_t$ that brings the population level from $x_t$ to $x_t - h_t$ and the effort $E_t$ needed to accomplish this result. We will a priori assume that there is a *marginal effort* involved, so that

$$E_t = \int_{x_t - h_t}^{x_t} \frac{1}{qy^b} dy \qquad (13)$$

for some $q$ and $b$. This is inspired by the fact that less effort is required when the stock is abundant, and can also be interpreted as an integral of infinitesimal Cobb-Douglas production functions (a standard economic model for productivity) where $b$ and $g$ are the corresponding elasticities. Estimated values obtained by least squares fitting are reported in table 1, while the resulting curve is compared with historical data in figure 1.

Costs involved in the Halibut fishery are divided into two categories: *fixed costs* and *variable costs*. Fixed costs include costs that are independent of the number and the duration of the trips a vessel makes (therefore generically independent from the effort $E_t$). For example, vessel repairs costs, license and insurance fees, mooring and dockage fees are typically considered fixed costs. We will denote with $K$ the sum of all the fixed costs, that will be incurred if and only if a harvest is undertaken.

Variable costs include all the expenses that are dependent on the effort level. Variable costs typically include fuel, maintenance, crew wages, gear repair and replacement. We assume that the total variable costs are proportional to the effort $E_t$ (measured in skate soaks) according to a constant $c$. Parameter $c$ is set to $200,000\$$ for 1000 skate soaks ($200\$$/skate) as estimated in [2]. Following the analysis of the historical variable and fixed costs for the halibut fishery carried on in [11], we assume $K = 5,000,000\$$ for area 3A. The unit price $p$ for the halibut is set to $4,300,000\$/10^6$ pounds, as in [2].

If we further assume a fixed discount rate $\delta = 0.05$, we obtain a formulation of management problem for the Halibut fishery in Area 3A that fits into the framework described in the previous section. In particular, the problem for an $N$ years horizon is that of finding an admissible policy $\pi = \{\mu_i\}_{i \in [1, N]}$ that maximizes the revenue $C_N^\pi(x)$ where $x_n$ is subject to (12), $h_n = \mu_n(x_n)$ and $R(x) = px - c\int_0^x \frac{1}{qy^b} dy$.

## 5.2 Optimal Policy

By using the dynamic programming approach on the problem discretized with a step size of $0.25 \times 10^6$ pounds, we compute the optimal policy for a management horizon of $N = 33$ years, that is the length of our original time series. As predicted by Theorem 1, the optimal policy $\pi^* = \{\mu_1, \ldots, \mu_N\}$ for the model we constructed for area 3A is a non stationary $(S - s)$ policy. In figure 2(a) we plot the function $\mu_1(\cdot)$ to be used in the first year (the values of $S_1$ and $s_1$ are 133 and 176.75 respectively). In words, the optimal policy dictates that at period $n$ a harvest is to be undertaken if and only if the current stock level is greater than $s_n$; in that case the stock is harvested down to $S_n$.

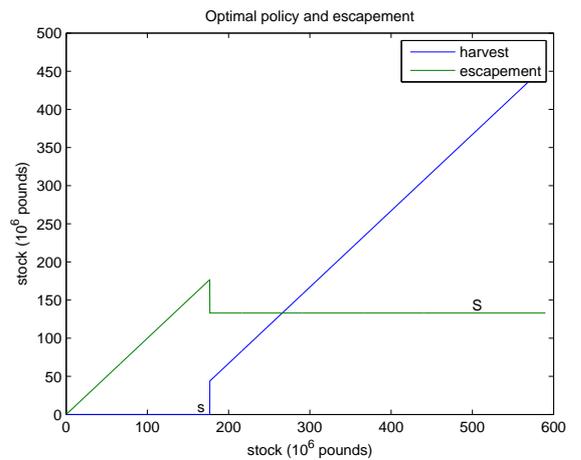

(a) Optimal rule for selecting harvests in the first year.

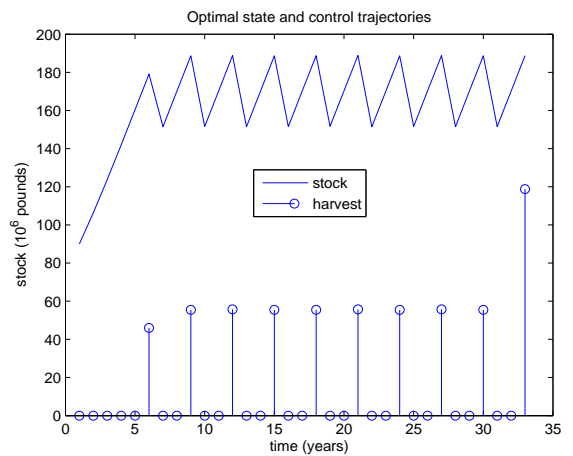

(b) Stock trajectory and corresponding optimal harvests.

Figure 2: The optimal policy.

The trajectory of the system when it is managed using the optimal policy is shown in figure 2, together with the corresponding optimal harvests. As we can see, the optimal policy is *pulsing*, in the sense that it involves periodic closures of the fishery, when no harvest should be undertaken so that the fish stock has time to recover. Of course, this kind of policy could be acceptable in practice only in combination with some rotation scheme among the different Areas, so that a constant yearly production can be sustained.

This scheme is very different from the Constant Proportional Policy (CPP) that has been traditionally used to manage the Halibut fishery. In fact a CPP works by choosing the yearly TAC as a fixed fraction of the current stock level $x_t$, and is aimed at maintaining the exploited stock size (the escapement) at a given fixed level. This policy can be seen as a simplified version of an $(S-s)$ policy where the two levels do not depend on the stage $n$ and coincide, thus defining the target stock size.

To see the advantage of the optimal $(S-s)$ policy, we compare it with the historical harvest proportions and with a CPP policy that uses the historical average harvest rate $a = 0.1277$. Table 2 summarizes the discounted revenues corresponding to an initial stock size $x_1 = 90.989$ million pounds, that is the estimated stock size in 1975.

| Policy | Disc. revenue ($) | Loss ($) |
|---|---|---|
| Optimal $S-s$ | $9.05141 \times 10^8$ | – |
| Historical rates | $7.06866 \times 10^8$ | $1.98275 \times 10^8$ |
| Average CPP | $6.51849 \times 10^8$ | $2.53292 \times 10^8$ |
| Rolling Horizon | $8.73605 \times 10^8$ | $3.1536 \times 10^7$ |

Table 2: Policy Comparison

Compared to the historical policy or the CPP policy, revenues for the optimal $(S-s)$ policy are about 35% higher, as reported in table 2. Notice that the comparison is done assuming a worst case realization of the stochasticity, or in other words that the nature is actively playing against the manager.

Notice that the large harvest prescribed by the optimal $(S-s)$ policy in the last year is an artifact of the finite horizon effect, caused by the fact that there is no reason not to exhaust the resource at the end of the management horizon (as long as it is profitable to harvest it). However it does not affect the comparison significantly due to the discount rate. In fact the (discounted) revenue for the entire last large harvest only accounts for less than 8% of the total revenue. This is confirmed by looking at the results obtained with a rolling horizon strategy that always picks the optimal action with a 33-years long management horizon in mind. As shown in figure 3, this (suboptimal) strategy is not affected by the finite horizon effect. The rolling horizon strategy still involves periodic closures of the fishery and significantly outperforms the historical policies, as reported in table 2.

To further clarify that the pulsing nature of the optimal harvests is not an artifact of the finite horizon, it is also interesting to notice that the theoretical results on the optimality of $(S-s)$ policies and the corresponding pulsing harvests can be carried over to the infinite horizon case via limiting arguments. The high level argument is that the *optimal value function* $C_n(x)$ converges uniformly to $C(x)$ as $n \to \infty$, while $P_n(x)$ converges uniformly to a function $P(x)$ as $n \to \infty$. Given that by Theorem 1 $P_n(x)$ is continuous and $K$-concave for all $n$, we have that $P(x)$ must be also continuous and $K$-concave. Using an argument similar to the one developed in section 3.3 and by using Lemma 1, one can show that there exists $S$ and $s$ such that the optimal stationary policy for the infinite horizon problem is an $(S-s)$ policy.

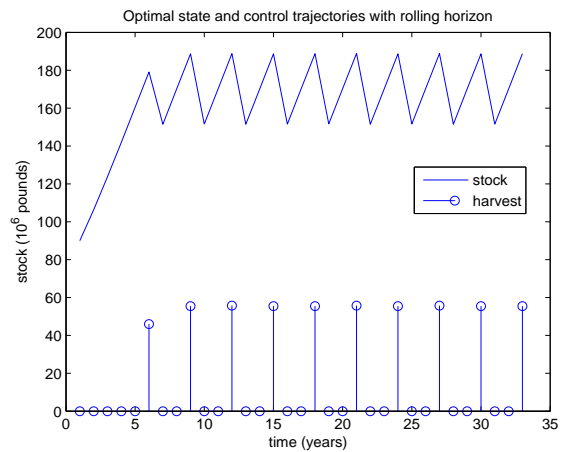

Figure 3: Harvests and stock trajectory with the rolling horizon strategy.

## 6 Conclusions

In this paper, we have analyzed the optimality of $(S-s)$ policies for a fairly general class of stochastic discrete-time resource allocation problems. When a non stationary $(S-s)$ policy is used, a harvest is undertaken at period $n$ if and only if the current stock level is greater than $s_n$; in that case the stock is harvested down to $S_n$. The framework developed is quite general and can be applied to problems arising in very different domains, such as natural resource management, crowdsourcing, pollution management. When assumptions of Theorem 1 are met, we have shown that there exists a non stationary $(S-s)$ policy that maximizes the utility in a worst case scenario.

A fundamental advantage of the game theoretic approach is that it completely avoids the problem of evaluating the probability distributions of the random variables describing the uncertainty affecting those systems, a task that is difficult or even impossible to accomplish in many practical circumstances. Given the consensus reached by the scientific community on the importance of understanding the role of uncertainty when dealing with renewable resources, we believe that worst-case scenario frameworks such as the one described here provide new insights and will become increasingly important.

To contribute to the effort of the Computational Sustainability community in tackling the fundamental sustainability challenges of our time, we consider an application of our model to a marine natural resource. This type of natural resources are in fact widely believed to be endangered due to over exploitation and generally poorly managed. Using Gulf of Alaska Pacific halibut data from the International Pacific halibut Commission (IPHC) annual reports, we formulated a real world case study problem that fits into our framework. In particular, our approach defines a policy with a guaranteed lower bound on the utility function that is structurally very different from the one currently employed.

As a future direction, we plan to study the effects of partial observability on the optimal policies by moving into a POMDP framework. Moreover, we aim at extending the results presented here to the multidimensional case by extending the theory on the so-called $(\sigma, S)$ policies from the inventory control literature.

# 7 Acknowledgments

This research is funded by NSF Expeditions in Computing grant 0832782.